\documentclass{svproc}

\usepackage{url}
\usepackage[numbers,sort&compress,sectionbib]{natbib}% Citation support using natbib.sty
\usepackage{color,soul}
\usepackage{tabularx}
\usepackage{soul}
\usepackage{graphicx}
\usepackage{amsmath}
\usepackage{caption}
\usepackage{subfig}
\usepackage{multicol}
\usepackage{multirow}
\usepackage{amssymb}
\usepackage{amsfonts}
\usepackage{graphicx}
\usepackage{float}
\usepackage{comment}
\usepackage{amsmath}
\usepackage{threeparttable}
\newcolumntype{Y}{>{\centering\arraybackslash}X}

\begin{document}
\mainmatter              % start of a contribution

\title{The impact of body and head dynamics on motion comfort assessment}
% %
\titlerunning{The impact of body and head dynamics}  % abbreviated title (for running head)
% %                                     also used for the TOC unless
% %                                     \toctitle is used
% %
\author{Georgios Papaioannou\inst{*} \and Raj Desai \and Riender Happee}
% %
\authorrunning{Georgios Papaioannou et al.} % abbreviated author list (for running head)
% %
% %%%% list of authors for the TOC (use if author list has to be modified)
\tocauthor{Georgios Papaioannou, Raj Desai, Riender Happee}
% %
\institute{Department Cognitive Robotics TU Delft, Delft, the Netherlands,\\
\email{g.papaioannou@tudelft.nl}}

\maketitle              % typeset the title of the contribution

\begin{abstract}

% Multiple theories exist to explain why motion sickness (MS) occurs. 
% According to the sensory conflict theory, MS is caused by the mismatch between sensed and anticipated sensory inputs.
% On the other hand, the postural instability theory attributes MS to prolonged uncoordinated configuration of the body and its segments, rather than sensory stimulation.
% During driving conditions, head  motion shows substantial rotations in response to seat excitations. In a range of experiments, it has been proven that head motion differs from vehicle motion and this is a key determinant of subjective and objective (dis)comfort. Meanwhile, state-of-the-art MS models focus mainly on the vestibular and visual conflicts, neglecting head rotation and body posture. 
Head motion is a key determinant of motion comfort and differs substantially from seat motion due to seat and body compliance and dynamic postural stabilization. This paper compares different human body model fidelities to transmit seat accelerations to the head for the assessment of motion comfort through simulations. Six-degree of freedom dynamics were analyzed using frequency response functions derived from an advanced human model (AHM), a computationally efficient human model (EHM) and experimental studies.
Simulations of dynamic driving show that human models strongly affected the predicted ride comfort (increased up to a factor 3). 
Furthermore, they modestly affected sickness using the available filters from the literature and ISO-2631 (increased up to 30$\%$), but more strongly affected sickness predicted by the subjective vertical conflict (SVC) model (increased up to 70$\%$).

%  Human body models of varying complexity will be coupled with available MS model to investigate their impact on MS occurrence through simulations. \dots
% We would like to encourage you to list your keywords within
% the abstract section using the \keywords{...} command.
\end{abstract}
\section{Introduction}

Automated vehicles (AVs) are considered one of the major automotive technological developments able to improve safety, environmental impact and accessibility. 
Key drivers for consumers to adopt them is the engagement in non-driving related tasks (NDRT). 
However, AVs envisaged designs are expected to provoke motion sickness (MS) in occupants jeopardizing their engagement in NDRT. 
There are multiple explanations for why MS occurs. 
According to the sensory conflict theory, MS is caused by the mismatch between sensed and anticipated (based on prior experience) sensory (visual, vestibular and somatosensory) inputs. 
On the other hand, the postural instability theory  attributes MS to the prolonged uncoordinated configuration of the body and its segments, rather than sensory stimulation. 
Head motion is a key determinant of (dis)comfort as perceived by sensory systems. 
A range of experiments has proven head translational motion to differ from vehicle motion with gains around one at low frequencies, oscillations at mid frequencies and with attenuation at higher frequencies. 
Moreover the head shows substantial rotations in response to seat translational excitation \cite{mirakhorlo2022effects,kato2006study,papaioannou2021assessment}. 
The transmission of motion from seat to head depends on seat compliance and on posture \cite{mirakhorlo2022effects}. 
Initial efforts to incorporate the effects of posture remain to be validated in particular in dynamic driving. At the same time, MS models that consider to some extent posture using seat-to-head transmissibility functions do not differentiate head from body posture.

This paper presents the coupling of different human body models to understand the body and head dynamics effect on the assessment of MS and ride comfort. 
More specifically, seat-to-head-transmissibilities from a set of experimental data and different human body model fidelities are coupled with an existing MS model as suggested by \cite{papaioannou2021assessment}. 
The models are used to answer the following research questions: (a) How do head and body dynamics impact the development of MS?, and (b) Could the use of a more detailed human body model enhance the MS models' prediction capabilities?
The human body models will include the active human body model from MADYMO \cite{tass2010madymo}, and a new 3D model \cite{desai2023evaluation}, also developed in MADYMO, which is faster than real-time offering efficient and robust exploration, while enabling for real-time control. 
Both models have been validated and are applied in combination with multibody models of compliant car seats.
% The paper is structured as follows: Section 2 presents the multidimensional human body models, and the metrics used to assess ride comfort and MS; Section 3 illustrates and discusses the results; Section 4 extracts conclusions and provides suggestions for further work.

\section{Methods $\&$ Materials}

% Add comment about interpolating data and initial frequency, and the new frequency.

\subsection{Multidimensional human body models}

\begin{figure}[ht!]
  \centering 
  \includegraphics[width=1\linewidth] {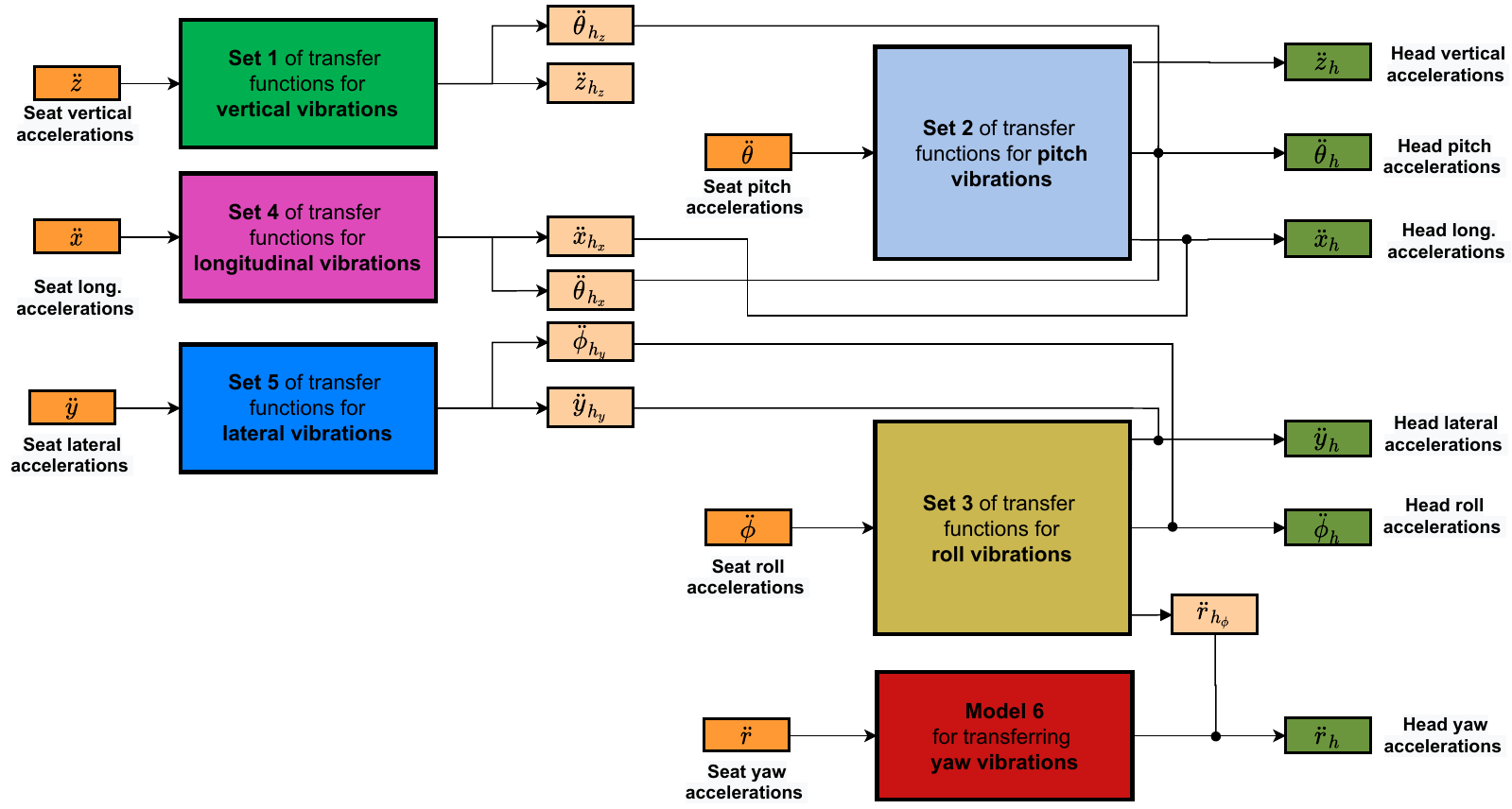}
  \caption{Multidimensional human body model, combining transfer functions based on experimental data or human body models of different fidelities.}
  \label{fig:human_body}
\end{figure}

The vehicle acceleration measurements ($\ddot{x}$, $\ddot{y}$, $\ddot{z}$, $\ddot{r}$, $\ddot{\phi}$ and $\ddot{\theta}$) extracted from IPG/CARMAKER simulations are transmitted to the occupants' head ($\ddot{x}_h$, $\ddot{y}_h$, $\ddot{z}_h$, $\ddot{r}_{h}$, $\ddot{\phi}_{h}$ and $\ddot{\theta}_{h}$) through frequency response functions (FRFs, gain and phases) extracted from different multi-dimensional models or experimental data sets. 
The FRFs are extracted with frequency step at 0.4 Hz, but they are interpolated at 0.005 Hz.
Vehicle motion data is Fourier transformed, the transfer functions are applied, data is transformed to the time domain with inverse Fourier, and comfort and sickness measures are derived. 
This is computationally much more efficient, as compared to using the full non-linear AHM and EHM models.
For this, the FRFs are extracted from different cases (i.e., simulation and experimental data) and they are grouped as six sets of transfer functions (Figure \ref{fig:human_body}, Set 1 - 6). 
In particular, the cases considered in this work consider transfer functions from the following simulation models or data:
\begin{itemize}
    \item \textbf{AHM}: The MADYMO detailed active human model (AHM) \cite{tass2010madymo} represents the 50th percentile male population and has been validated for impact conditions, vibration and dynamic driving \cite{mirakhorlo2022simulating}. 
    The AHM consists of 190 bodies (182 rigid bodies and 8 flexible bodies) and finite element surfaces capture the skin for contact interaction.
    \item \textbf{EHM}: An efficient human model (EHM) for comfort analysis \cite{desai2023evaluation}. 
    The EHM is also built in MADYMO and considers a functional set of body segments, and joint degrees of freedom, with only those that have a significant impact on the kinematics and dynamics of the body. 
    The model has proven to provide results close to experimental data and the AHM. 
    \item \textbf{EXP}: A combination of published experimental data \cite{Paddan1994,Paddan1999,mirakhorlo2022effects} that studied the occupants' seat-to-head transmissibility (STHT).
\end{itemize}

The six sets of transfer functions (Figure \ref{fig:human_body}) employed are applied as follows. 
Regarding the head vertical accelerations, Set 1 includes two transfer functions ($T_{z_{J}}$ and $T_{z_{\theta,J}}$), that transmit the seat vertical acceleration to the occupants' head and evaluate the pitch head accelerations ($\ddot{\theta}_{h_z}$) that are induced by the seat vertical accelerations. 
This process is conducted through Equation \ref{eq:Tver}:

\begin{equation}
\begin{split}
   \ddot{z}_{h_z}     ~ = &  ~T_{z_{J}}        ~~ \ddot{z}\\
   \ddot{\theta}_{h_z} ~ = &  ~T_{z_{\theta,J}} ~ \ddot{z}
  \label{eq:Tver}
\end{split}
\end{equation}

\noindent where $\ddot{z}_{h_z}$ is the head vertical acceleration without considering the impact of any other excitations on it;
$\ddot{\theta}_{h_z}$ are the pitch head accelerations that are induced by the seat vertical acceleration; 
$J$ corresponds to source of the data, i.e., recorded head motion from experiments ($J$=EXP) \cite{mirakhorlo2022effects}, the active human model from MADYMO ($J$=AHM) \cite{mirakhorlo2022simulating}, the efficient human model ($J$=EHM), and no human model ($J$=NHM) which corresponds to $T_{z_{NHM}}$=1 assuming that the head motion equals the seat motion. 

Regarding the transmissibility of pitch ($\ddot{\theta}$) and roll ($\ddot{\phi}$) accelerations, multidimensional transfer functions (Set 2 and 3) are designed based on the literature \cite{Paddan1994}. %, transferring the corresponding rotational accelerations to the head and considering its effect to the other translational or rotational accelerations.
In particular, the pitch transmissibility model ($\ddot{\theta}$, Set 2) includes three transfer functions. 
The first, $T_{\theta}$ is used to transmit the seat pitch accelerations to the head, while the other two ($T_{\theta_x}$ and $T_{\theta_z}$) consider the effect of seat pitch accelerations to the head vertical ($\ddot{z}_{h_\theta}$) and longitudinal accelerations $(\ddot{x}_{h_\theta})$.
These transfer functions are used according to Equation \ref{eq:Ttheta1}.

\begin{equation}
\begin{split}
   \ddot{x}_{h_\theta}        ~ = &  ~T_{\theta_x} \ddot{\theta} \\
   \ddot{z}_{h_\theta}        ~ = &  ~T_{\theta_z} \ddot{\theta} \\
   \ddot{\theta}_{h_\theta}   ~ = &  ~T_{\theta} ~ \ddot{\theta}
\end{split}
\label{eq:Ttheta1}
\end{equation}

Similarly with pitch, the set of transfer function for roll ($\ddot{\phi}$, Set 3) also includes three transfer functions \cite{Paddan1994}. 
The first, $T_{\phi}$ is used to transmit the seat roll accelerations to the head ($\ddot{\phi}_{h_{\phi}}$), while the other two ($T_{\phi_y}$ and $T_{\phi_R}$), consider the effect of seat roll accelerations to the head lateral ($\ddot{y}_{h_{\phi}}$) and yaw ($\ddot{r}_{h_{\phi}}$) accelerations.
These are used according to Equation \ref{eq:Trol}:

\begin{equation}
\begin{split}
   \ddot{y}_{h_{\phi}}    ~ = &  ~T_{\phi_y} \ddot{\phi}\\
   \ddot{r}_{h_{\phi}}    ~ = &  ~T_{\phi_r} \ddot{\phi} \\
   \ddot{\phi}_{h_{\phi}} ~ = &  ~T_{\phi} ~ \ddot{\phi}
\end{split}   
\label{eq:Trol}
\end{equation}

Two more sets of transfer functions  (Sets 4 and 5) are developed to transmit the longitudinal and lateral accelerations to the head including their interactions with other rotational or translational accelerations. 
Regarding the longitudinal accelerations ($\ddot{x}$, Set 4), $T_{x_{J}}$ transmits the seat longitudinal accelerations to the occupants' head ($\ddot{x}_{h_x}$), while $T_{x_{\theta,J}}$ evaluates the pitch head accelerations ($\ddot{\theta}_{h_x}$) that are induced by the seat longitudinal accelerations:

\begin{equation}
\begin{split}
    \ddot{x}_{h_x}      ~ = &  ~T_{x,J}   ~ \ddot{x} \\
    \ddot{\theta}_{h_x} ~ = &  ~T_{x_{\theta,J}}  \ddot{x}
\end{split}
\label{eq:Tlon}
\end{equation}

\noindent For the lateral accelerations ($\ddot{y}$, Set 5), $T_{y_{J}}$ transmits the seat lateral accelerations to the occupants' head, while $T_{y_{\phi,J}}$ and $T_{y_{r,J}}$ evaluate the roll ($\ddot{\phi}_{h_y}$) and yaw ($\ddot{r}_{h_y}$) head accelerations that are induced by the seat lateral accelerations:

\begin{equation}
\begin{split}
   \ddot{y}_{h_y}    ~ = &  ~T_{y,J}       ~~ \ddot{y}\\
   \ddot{r}_{h_y}    ~ = &  ~T_{y_{r,J}}    ~ \ddot{y} \\
   \ddot{\phi}_{h_y} ~ = &  ~T_{y_{\phi,J}} ~ \ddot{y}
\end{split}   
\label{eq:Tlat}
\end{equation}

\noindent Finally, the yaw accelerations ($\ddot{r}$, Set 6) are transmitted from the seat to the head ($T_{yaw}$) according to an experimental transfer function \cite{Paddan1999}:

\begin{equation}
   \ddot{r}_{h_r}  =  T_{r} \ddot{r}
  \label{eq:Tyaw}
\end{equation}

The total accelerations including the above calculated interactions are calculated as follows: 

\begin{equation}
\begin{split}
   \ddot{z}_{h}      ~ = &  ~\ddot{z}_{h_z} + \ddot{z}_{h_\theta} \\
   \ddot{x}_{h}      ~ = &  ~\ddot{x}_{h_x} + \ddot{x}_{h_\theta} \\
   \ddot{y}_{h}      ~ = &  ~\ddot{y}_{h_y} + \ddot{y}_{h_\phi}  \\
   \ddot{\phi}_{h}   ~ = &  ~\ddot{\phi}_{h_\phi} + \ddot{\phi}_{h_y} \\
   \ddot{\theta}_{h} ~ = &  ~\ddot{\theta}_{h_\theta} + \ddot{\theta}_{h_z} + \ddot{\theta}_{h_x}\\
   \ddot{r}_{h}      ~ = &  ~\ddot{r}_{h_r} + \ddot{r}_{h_y} + \ddot{r}_{h_\phi}  \\
\end{split}   
\label{eq:Ttot}
\end{equation}

% As mentioned before, the multidimensional human body model (Figure \ref{fig:human_body}) combines transfer functions obtained from experimental data \cite{Paddan1999,mirakhorlo2022effects} and human models of different fidelity. 
As mentioned before, to answer the research questions of this work, four multi-dimensional human body models are developed using different combinations of transfer functions. 
These combinations are illustrated in Table \ref{tab:HM_comb}.
For the first three multi-dimensional human body models (EXP, AHM and EHM), Set 1, 4 and 5 are extracted from published experimental data (EXP) the literature \cite{mirakhorlo2022effects} and different human body model fidelity (AHM, EHM and NHM) \cite{mirakhorlo2022simulating,desai2023evaluation}.
Set 2,3 and 6 are the same for configurations AHM, EHM, EXP and extracted from published experimental data \cite{Paddan1994,Paddan1999}.  
For NHM, all transfer functions are equal to 1, i.e., seat accelerations are transferred to the occupant's head as is.

\begin{table}[h!]
    \centering
    \caption{Combinations of transfer functions to extract different configurations for different multidimensional human body models.}
    \label{tab:HM_comb}  
    \begin{tabularx}{\linewidth}{|c|*{3}{Y|}c|}
        \hline
                          & \textbf{EXP}   & \textbf{AHM}      & \textbf{EHM}     & \textbf{NHM}    \\
        \hline
        \textbf{Set 1}  & $T_{z_{,EXP}}$, $T_{z_{\theta,EXP}}$ \cite{mirakhorlo2022effects}  & $T_{z_{,AHM}}$, $T_{z_{\theta,AHM}}$ \cite{mirakhorlo2022simulating} & $T_{z_{,EHM}}$, $T_{z_{\theta,EHM}}$ \cite{desai2023evaluation}  & \multirow{6}{*}{1}  \\
        \cline{1-4}
        \textbf{Set 2}  & \multicolumn{3}{c|}{$T_{\theta_x}$, $T_{\theta_z}$, $T_{\theta}$ \cite{Paddan1994}}  & \\
        \cline{1-4}
        \textbf{Set 3}  & \multicolumn{3}{c|}{$T_{\phi_y}$, $T_{\phi_r}$, $T_{\phi}$ \cite{Paddan1994}}  & \\
        \cline{1-4}
        \textbf{Set 4}  & $T_{x,EXP}$, $T_{x_{\theta,EXP}}$  \cite{mirakhorlo2022effects}   & $T_{x,AHM}$, $T_{x_{\theta,AHM}}$ \cite{mirakhorlo2022simulating}   & $T_{x,EHM}$, $T_{x_{\theta,EHM}}$ \cite{desai2023evaluation} &    \\
        \cline{1-4}
        \textbf{Set 5}  & $T_{y_{,EXP}}$, $T_{y_{r,EXP}}$, $T_{y_{\phi,EXP}}$ \cite{mirakhorlo2022effects}  & $T_{y_{,AHM}}$, $T_{y_{r,AHM}}$, $T_{y_{\phi,AHM}}$ \cite{mirakhorlo2022simulating}    & $T_{y_{,EHM}}$, $T_{y_{r,EHM}}$, $T_{y_{\phi,EHM}}$ \cite{desai2023evaluation}  &  \\
        \cline{1-4}
        \textbf{Set 6}  & \multicolumn{3}{c|}{$T_{r}$ \cite{Paddan1999}} & \\
        \hline
    \end{tabularx}
\end{table}

\subsection{Motion comfort assessment}

ISO-2631:1998 \cite{ISO2631} provides objective guidelines for measurement and evaluation of human exposure to whole-body mechanical vibration and repeated shock. 
According to the guidelines, two comfort metrics are derived: (1) Ride Comfort (RC) emphasizing the higher frequencies (mainly above 1 Hz); (2) Motion Sickness (MS) emphasizing the lower frequencies (mainly below 1 Hz). 
Both metrics apply frequency weighting to six degrees of freedom motion including three dimensional translation and  rotation the head.
% The Ride Comfort (RC) is expected to capture general motion (dis)comfort due to vibration and abrupt motion and is deemed relevant to active motion (driving) and passive motion (being driven).
% The second measure is suitable for passive motion (being driven).
Additionally, for a more in depth analysis of MS, we employ the subjective vertical sensory conflict model (SVC) to assess the motion sickness incidence in various paths \cite{inoue2023revisiting}. 

According to the standard, comfort is assessed by combining the root mean square (RMS) values of weighted accelerations ($RC_{W_{i}}$), translational and rotational, measured at the vehicle's centre of gravity.
More specifically, the RMS value of each acceleration is calculated as follows: 

\begin{equation}
\label{eq:accel}
	RC_{W_{i}}=  \bigg( \frac{1}{t}  \int_{0}^{t} a_{i_W}^2 d\tau \bigg)^{\frac{1}{2}}
\end{equation}

\noindent where $i$ is the acceleration type, either translational ($\ddot{x}$, $\ddot{y}$ and $\ddot{z}$) or rotational ($i$= $rx$ for $\ddot{\phi}$, $ry$ for $\ddot{\theta}$ and $rz$ for $\ddot{r}$) as defined in the standard \cite{ISO2631}, while $a_{W_i}$ stands for the weighted accelerations in the time domain.
After multiplying each of the $RC_{iW_{rms}}$ by appropriate factors ($k_i$), they are all summed and the overall comfort metric is calculated: 

\begin{equation}
\label{eq:RC}
RC = \bigg( \sum_{i=1}^{6} k_i^2 RC_{W_{i}}^2 \bigg)^{1/2}
\end{equation}

\noindent where $k_i$ is the multiplying factor for each term ($i$=$x$, $y$, $z$, $rx$, $ry$ and $rz$) which can be found in ISO-2631 \cite{ISO2631}.
Equation \ref{eq:RC} is used with two different sets of weighting filters for the translational and rotational accelerations to objectively assess RC and MS. 
%  (Table \ref{tab:filters}).
For RC, $WP_k$ and $WP_e$ are applied for the vertical and all the rotational accelerations, respectively.
No filter is used in x and y direction according to the standard \cite{ISO2631}.
For MS, $WP_{f_x}$ \cite{Griffin2002}, $WP_{f_y}$ \cite{Donohew2004} and $WP_{f}$ \cite{ISO2631} are used for the x, y and z direction, while $WA_{f_r}$ is used for all the rotational vibrations \cite{Howarth2003}.
More details can be found in literature \cite{papaioannou2023motion}.

% %%%%%%%%%%%%%%%%%%%%%%%%%%%%%%%%%%%%%%%%%%%%%%%%%%%%%%%%%%%%%%%%%%%%%%%%%%%%%%%%%%%%%%%%

% \begin{table}[h!]
%     \centering
%     \caption{Applied weighting filters in translational and rotational accelerations for the assessment of ride comfort and motion sickness.}
%     \label{tab:filters}    
%     % \begin{tabular}{|c|c|c|c|c|c|c|}
%     \begin{tabularx}{\linewidth}{|c|*{6}{Y|}}
%          % \hline
%          % \multicolumn{7}{|c|}{\multirow{2}{*}{\textbf{Weighting Filters}}} \\
%          % \multicolumn{7}{|c|}{} \\         
%          \hline
%          \multirow{2}{*}{\textbf{Application}}  & \multicolumn{6}{c|}{\textbf{Vibration}}\\
%          \cline{2-7} 
%                                  &  \textbf{x}  & \textbf{y}  & \textbf{z}  &  \textbf{rx}  & \textbf{ry}  & \textbf{rz} \\     
%          \hline
%          Ride comfort (RC)       & - & - & $WP_k$~~\cite{ISO2631}  & \multicolumn{3}{c|}{$WA_e$ \cite{ISO2631}}  \\ 
%          \hline
%          Motion sickness (MS)    &  $WP_{f_x}$ \cite{Griffin2002}   & $WP_{f_y}$  \cite{Donohew2004} & $WP_{f_z}$ \cite{ISO2631}  & \multicolumn{3}{c|}{$WA_{f_r}$\cite{Howarth2003}} \\ 
%          \hline
%     \end{tabularx}
% \end{table}

% %%%%%%%%%%%%%%%%%%%%%%%%%%%%%%%%%%%%%%%%%%%%%%%%%%%%%%%%%%%%%%%%%%%%%%%%%%%%%%%

\section{Results}

\begin{figure}[!ht]
\centering

  \begin{minipage}[b]{.49 \linewidth}
  	 \includegraphics[width=\linewidth]{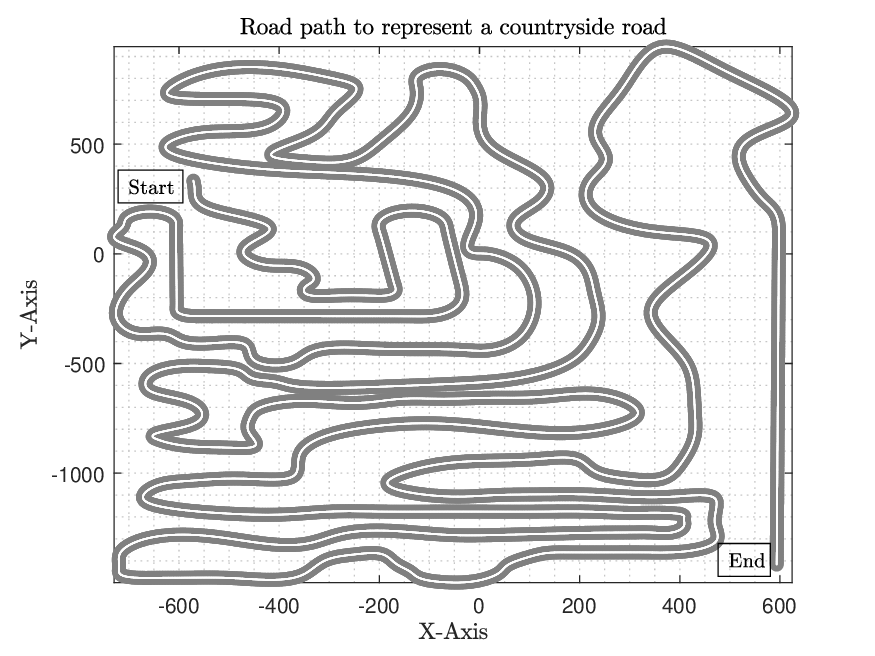}
     \begin{center}
     \vspace{-0.4 cm}
     (a)
     \end{center}
  \end{minipage}
  \centering
  \begin{minipage}[b]{.49 \linewidth}
  	 \includegraphics[width=\linewidth]{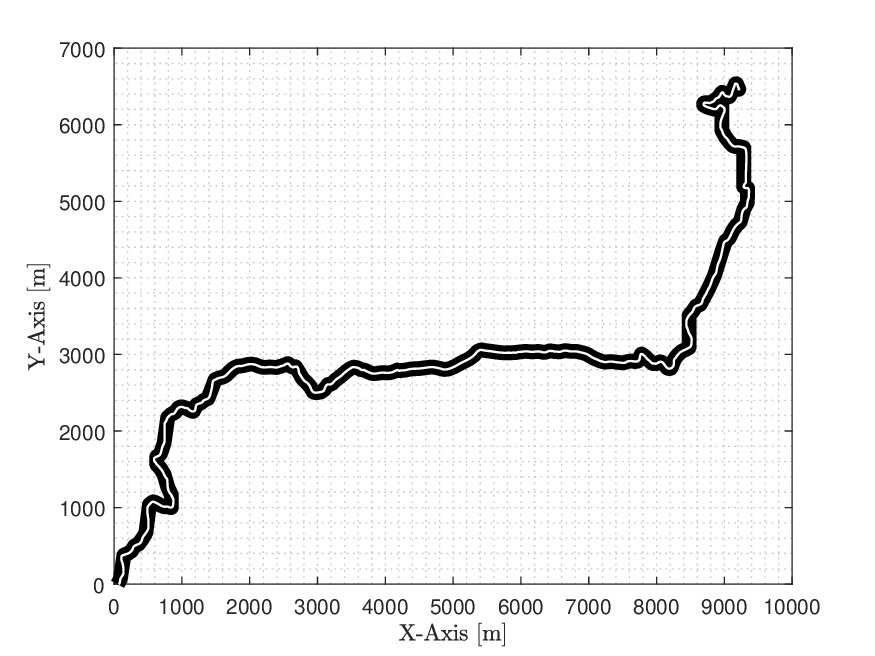}
     \begin{center}
     \vspace{-0.4 cm}
     (b)
     \end{center}
  \end{minipage}  
\caption{(a) Path 1 and (b) Path 2 trajectories (X-Y) \cite{inoue2023revisiting}.}  
\label{fig:paths}
\end{figure}

\begin{figure}[!ht]
\centering

  \begin{minipage}[b]{.49 \linewidth}
  	 \includegraphics[width=\linewidth]{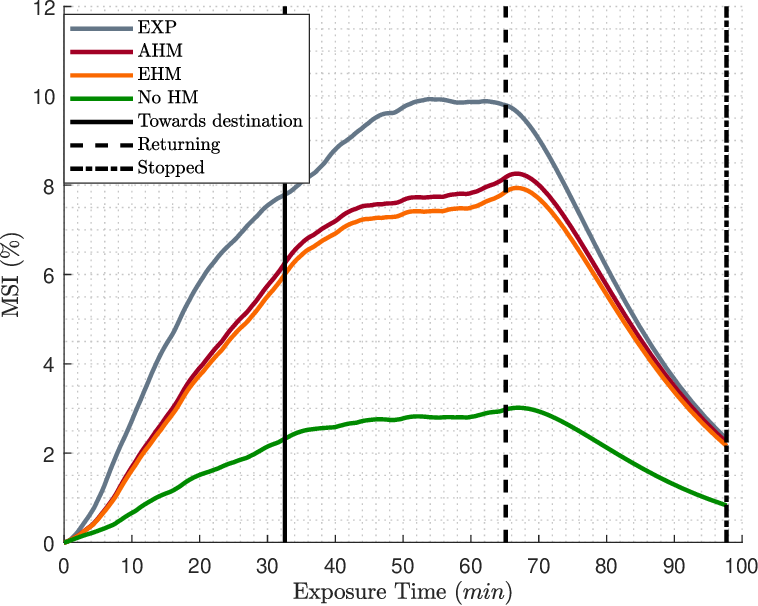}
     \begin{center}
     \vspace{-0.4 cm}
     (a)
     \end{center}
  \end{minipage}
  \centering
  \begin{minipage}[b]{.49 \linewidth}
  	 \includegraphics[width=\linewidth]{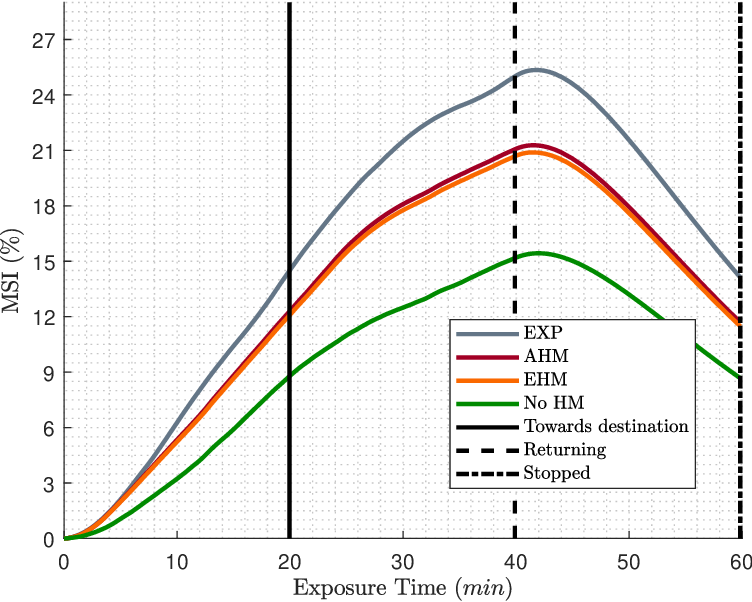}
     \begin{center}
     \vspace{-0.4 cm}
     (b)
     \end{center}
  \end{minipage}  
\caption{Motion sickness incidence for (a) Path 1 and (b) Path 2 as calculated by the subjective vertical sensory conflict model \cite{inoue2023revisiting}.}  
\label{fig:MSI}
\end{figure}

The impact of body and head dynamics on motion comfort assessment is investigated for two different road paths: (a) Path 1: an artificial road path designed using IPG/CARMAKER ($\sim$ 21.1 km, Figure \ref{fig:paths}a) \cite{papaioannou2021assessment}, (b) Path 2: a real countryside road path from Chrysovitsi to Stemnitsa in Arcadia, Greece ($\sim$ 17.0 km, Figure \ref{fig:paths}b) \cite{papaioannou2022k}.
The path characteristics and driving dynamics are illustrated in Table \ref{tab:path_vehicle}, and were extracted through simulations in IPG/CARMAKER with the IPG Driver.
The analysis includes the assessment of RC (Table \ref{tab:RC_assess}) per translational and rotational acceleration (RC$_i$, where $i$=x, y, z, rx, ry, and rz) and in total (RC$_{total}$). 
Similar, calculations are presented for MS in Table \ref{tab:RC_assess}. 
Furthermore, for a deeper analysis of MS, the motion sickness incidence (MSI) is calculated through a validated MS model \cite{inoue2023revisiting} (\ref{fig:MSI}).

\begin{table}[h!]
    \centering
    \caption{Path characteristics $\&$ vehicle driving dynamics (mean $\pm$ deviation).}
    \label{tab:path_vehicle}    
    % \begin{tabular}{|c|c|c|c|c|c|c|}
    \begin{tabularx}{\linewidth}{|c|c|c|*{4}{Y|}}
         \hline
                                                 & Path     & Road     & \multirow{2}{*}{$V$ (m/s)}         &  \multirow{2}{*}{$a_x$ (m/s$^2$)}    &  \multirow{2}{*}{$a_y$ (m/s$^2$)}   &  \multirow{2}{*}{$a_z$ (m/s$^2$)} \\
                                                 & distance (km) & profile  & & & & \\
         \hline
         \textbf{Path 1} & 21.1 & Class B & 10.72 $\pm$ 0.96  &  0.01 $\pm$ 0.32    &  0.01 $\pm$ 0.98   &  0.01 $\pm$ 1.55 \\
         \hline
         \textbf{Path 2}  & 17.0 & Class C & 11.04 $\pm$ 2.91  &  0.04 $\pm$ 0.70    &  0.01 $\pm$ 1.24   &  0.01 $\pm$ 2.05 \\
         \hline 
    \end{tabularx}
\end{table}

According to Table \ref{tab:RC_assess}, the two human body models (EHM and AHM) and the experimental dynamics (EXP) provide similar comfort predictions in both paths but differ strongly from results ignoring the dynamics of seat to head transmission (NHM). 
More specifically, EHM and AHM show minor differences ($\sim$ 5$\%$). 
On the other hand, NHM provides an underestimation of the RC$_{total}$ in Path 1 ranging from 46$\%$ to 62$\%$ compared to EHM and EXP, which provide the lowest and the highest RC estimation respectively. 
Greater underestimation takes place in Path 2 (62$\%$ to 68$\%$). 
The source of these great differences compared to NHM can be identified in the translational or rotational acceleration (RC$_j$) comfort calculation for both paths. 
In particular, the impact of the longitudinal acceleration on RC (RC$_x$) is increased by 5-6 times compared to the NHM, while the one of the lateral (RC$_y$) and vertical (RC$_z$) by 1.5 - 2 times. 
However, RC$_y$ impact is insignificant to RC$_{total}$, even after such increases. 
Similarly, the impact of rotational accelerations has also increased compared to NHM. 
More specifically, the roll ($ry$) and yaw ($rz$) RC metrics are increased by 4-5 times, but their impact compared to the RC$_{total}$ is negligible. 
On the other hand, the RC$_{rx}$ (ride comfort provoked by the pitch accelerations) is increased by more than 10 times, greatly increasing their impact on the overall ride comfort assessment.

As far as the MS assessment is concerned, the MS$_{total}$ does not illustrate any significant difference between the human body models considered (EXP, AHM and EHM) to transfer the seat accelerations to the head. 
However, there is significant underestimation ($\sim$ 32$\%$) in MS$_{total}$ when no human model is used (NHM) compared to the human body models (EXP, AHM and EHM).
Furthermore, albeit MS$_{total}$ being unaffected by the different human models, differences ($\sim$ 8-15 times larger for Path 1 and 2) can be identified to the MS$_{ry}$, which is the motion sickness provoked by the head pitch accelerations explicitly. 
However, these differences hardly affect the MS$_{total}$, since the weighting ($k_{ry}$) suggested by the ISO standards diminishes their impact. 
Similarly with the RC assessment, AHM and EHM provide similar assessments for MS$_{total}$ and for each MS$_{i}$.
This remark further validates the fact that the EHM provides accurate motion comfort assessment results compared to the more detailed and time consuming AHM, being a promising solution for real time control applications. 

On contrary with the traditional ISO-2631 based MS assessment (Table \ref{tab:MS_assess}) where MS$_{total}$ illustrate negligible differences between the different multi-dimensional human body models, the MSI as evaluated by the sensory conflict model \cite{inoue2023revisiting} illustrates significant changes (Figure \ref{fig:MSI}). 
According to Figure \ref{fig:MSI}, the NHM underestimated MSI by 70$\%$ and 40$\%$ compared to EXP in Path 1 and 2, respectively.
Meanwhile, the AHM and EHM provided similar responses with minor differences, which is in alignment with the previous remark regarding EHM's capabilities. 
However, AHM and EHM still present 15$\%$ and 19$\%$ less MSI than EXP. 
The source of these differences in MSI could be related with the difference identified in MS$_{ry}$ (Table \ref{tab:MS_assess}), since no other MS$_i$ illustrates any significant difference among the different human body model fidelities. 
This is in alignment with previous literature where head rotations are a key determinant of (dis)comfort and motion sickness as perceived by the sensory system. 

\begin{table}[h!]
    \centering
    \caption{Ride comfort assessment in Path 1 and 2 using different human body model fidelities. RC is calculated per translational and rotational acceleration (RC$_i$, where $i$=x, y, z, rx, ry, and rz) and in total (RC$_{total}$).}
    \label{tab:RC_assess}  
    \begin{tabularx}{\linewidth}{|c|*{8}{Y|}}

        \hline
                              & \multicolumn{4}{c|}{\textbf{Path 1}} & \multicolumn{4}{c|}{\textbf{Path 2}} \\
        \hline
                              & \textbf{EXP}     & \textbf{AHM}     & \textbf{EHM}    & \textbf{NHM}    & \textbf{EXP}     & \textbf{AHM}     & \textbf{EHM}    & \textbf{NHM} \\
        \hline
        \textbf{RC}$_x$       & 0.536 & 0.540 & 0.540 & 0.102 & 1.520  & 1.532 & 1.531 & 0.240 \\
        \hline
        \textbf{RC}$_y$       & 0.070 & 0.067 & 0.071 & 0.054 & 0.305  & 0.293 & 0.308 & 0.232 \\
        \hline
        \textbf{RC}$_z$       & 2.717 & 2.062 & 1.999 & 1.407 & 3.650  & 2.866 & 2.744 & 1.719 \\
        \hline
        \textbf{RC}$_{rx}$    & 0.169 & 0.136 & 0.138 & 0.041 & 0.803  & 0.555 & 0.588 & 0.151 \\
        \hline
        \textbf{RC}$_{ry}$    & 6.099 & 4.481 & 4.030 & 0.406 & 10.623 & 9.260 & 8.779 & 0.985 \\
        \hline
        \textbf{RC}$_{rz}$    & 0.090 & 0.073 & 0.085 & 0.021 & 0.432  & 0.348 & 0.347 & 0.051 \\
        \hline
        \textbf{RC}$_{total}$ & 3.693 & 2.788 & 2.626 & 1.421 & 5.835  & 4.949 & 4.738 & 1.798 \\
        \hline
    \end{tabularx}
\end{table}

\section{Conclusions}

To sum up, this paper investigated the impact of head and body dynamics in the assessment of motion comfort (i.e., motion sickness and ride comfort) using objective metrics that exist in the literature and various human body model fidelities to transfer seat-to-head accelerations. Advanced and detailed human body models, which can be time consuming, are not required to assess motion comfort, and the efficient human models are promising and capable to provide accurate results with similar prediction capabilities in real time applications. 

\begin{table}[h!]
    \centering
    \caption{Motion sickness assessment in Path 1 and 2 using different human body model fidelities, derived using ISO (for MSI see Fig 3). MS is calculated per translational and rotational acceleration (MS$_j$, where $i$=x, y, z, rx, ry, and rz) and in total (MS$_{total}$).}
    \label{tab:MS_assess}  
    \begin{tabularx}{\linewidth}{|c|*{8}{Y|}}

        \hline
                              & \multicolumn{4}{c|}{\textbf{Path 1}} & \multicolumn{4}{c|}{\textbf{Path 2}} \\
        \hline
                              & \textbf{EXP}     & \textbf{AHM}     & \textbf{EHM}    & \textbf{NHM}    & \textbf{EXP}     & \textbf{AHM}     & \textbf{EHM}    & \textbf{NHM} \\
        \hline
        \textbf{MS}$_x$       & 0.162 & 0.162 & 0.162 & 0.114 & 0.387 & 0.388 & 0.389 & 0.271 \\
        \hline
        \textbf{MS}$_y$       & 1.011 & 1.011 & 1.011 & 0.695 & 1.540 & 1.540 & 1.540 & 1.072 \\
        \hline
        \textbf{MS}$_z$       & 0.032 & 0.032 & 0.032 & 0.032 & 0.061 & 0.061 & 0.061 & 0.059 \\
        \hline
        \textbf{MS}$_{rx}$    & 0.007 & 0.006 & 0.006 & 0.004 & 0.021 & 0.015 & 0.016 & 0.007 \\
        \hline
        \textbf{MS}$_{ry}$    & 0.301 & 0.184 & 0.171 & 0.020 & 0.391 & 0.266 & 0.244 & 0.026 \\
        \hline
        \textbf{MS}$_{rz}$    & 0.624 & 0.624 & 0.624 & 0.027 & 0.816 & 0.816 & 0.816 & 0.046 \\
        \hline
        \textbf{MS}$_{total}$ & 1.039 & 1.035 & 1.034 & 0.705 & 1.605 & 1.601 & 1.601 & 1.107 \\
        \hline
    \end{tabularx}
\end{table}

The main contributions can be summarized as follows. 
Transfer functions (6DOF) capturing the seat to head transmission (STHT) were used to derive head motion as function of vehicle motion. 
Three classes of transfer functions based on full body models (AHM and EHM) and experimental data (EXP) were compared.
The AHM is computationally complex and takes around 100 times real time, whereas the EHM is faster than real time when using a relatively simple seat model.
In this paper, we derived 6D transfer functions from the AHM and EHM providing a further speedup to more than 6000-8000 times faster than real time. 
For instance, we simulate $\sim$ 1965 s in 0.32 s or 19807s in 0.25 s depending the path. 
This creates highly efficient solutions usable in model predictive control for automated vehicle motion control and motion cuing in driving simulators.
Including STHT strongly affected ride comfort (increased up to a factor 3) and modestly affected sickness using the traditional ISO-2631 filters (increased up to 30$\%$) but more strongly affected sickness predicted by the subjective vertical conflict (SVC) model (increased up to 70$\%$).
More specifically, the three different human body model fidelities led to more than 50$\%$ differences in overall ride comfort assessment regardless of the path. 
The impact of head rotations (pitch, roll and yaw) on ride comfort is diminished when no human model is used, i.e., assuming that the head accelerations are equal to the seat accelerations.
The head pitch rotations are a key determinant of motion sickness as perceived by the sensory system, but standardized metrics are unable to capture this impact. 

Further work is in progress comparing results with subjective assessment, investigating how different human body models can affect comfort and if models should be tuned differently taking into account posture and seat. \newline 

\noindent \textbf{Acknowledgement:} We acknowledge the support of Toyota Motor Corporation in funding the contribution of Raj Desai.

\bibliographystyle{spbasic}
\bibliography{IAVSD23_Postural_Stability_Papaioannou}

\end{document}